\documentclass[journal,twoside,web]{ieeecolor}
\usepackage{tmi}
\usepackage{cite}
\usepackage{amsmath,amssymb,amsfonts}
\usepackage[mathscr]{eucal}
\usepackage{graphicx}
\usepackage{textcomp}
\usepackage{float}
\usepackage{subfig}
\usepackage{url}
\usepackage{xcolor}
\usepackage{hyperref}
\usepackage{multirow}
\usepackage{longtable}
\usepackage{booktabs}
\usepackage{threeparttable}
\usepackage{algorithmicx}
\usepackage[linesnumbered,ruled,vlined]{algorithm2e}
\usepackage{algpseudocode}

\newcommand{\std}[1]{\,$\scriptstyle\pm\,\scriptstyle{#1}$}

\newcommand{\best}[1]{\textcolor{red}{\textbf{#1}}}
\newcommand{\second}[1]{\textcolor{blue}{\textbf{#1}}}
\hypersetup{hidelinks,
	colorlinks=true,
	allcolors=blue,
	pdfstartview=Fit,
	breaklinks=true}

\def\BibTeX{{\rm B\kern-.05em{\sc i\kern-.025em b}\kern-.08em
	T\kern-.1667em\lower.7ex\hbox{E}\kern-.125emX}}

\begin{document}

\title{CNsEMD: An Expert-Annotated Multi-Field-Strength MRI Dataset and a Hyperspherical Manifold Network for Multimodal Cranial Nerve Parcellation}

\author{Lei~Xie, Junxiong~Huang, Guoqiang~Xie, Jiawei~Zhang,
	Jiahao~Huang, Tianling~Lyu, Ye~Wu, Mingchu~Li, Shoujun Yu, Shanshan~Wang, Qingrun~Zeng, and Yuanjing~Feng
	\thanks{This work was sponsored in part by the Natural Science Foundation of Zhejiang Province under Grant No. LMS25F030004; National Natural Science Foundation of China under Grants No. U22A2040, U23A20334 and 62403428. (\textit{Corresponding author: Yuanjing~Feng (fyjing@zjut.edu.cn).})}
	\thanks{Lei~Xie, Tianling Lyu, Qingrun~Zeng, and Yuanjing~Feng are with the Advanced Interdisciplinary Science and Technology, Zhejiang University of Technology, Hangzhou 310014, China.}
	\thanks{Junxiong Huang, Jiawei Zhang, and Jiahao~Huang are with the College of Information Engineering, Zhejiang University of Technology, Hangzhou 310023, China.}
	\thanks{Ye Wu is with School of Computer Science and Technology, Nanjing University of Science and Technology, Nanjing 210094, China.}
	\thanks{Mingchu Li is with the Department of Neurosurgery, Capital Medical University Xuanwu Hospital, Beijing 100053, China.}
	\thanks{Shoujun Yu and Shanshan Wang is with the Paul C. Lauterbur Research Center for Biomedical Imaging, Shenzhen Institutes of Advanced Technology, Chinese Academy of Sciences, Shenzhen 518055, China.}
	\thanks{Guoqiang~Xie is with the Department of Neurosurgery, Nuclear Industry 215 Hospital of Shaanxi Province, Xianyang 712000, China.}
}

\maketitle

\begin{abstract}
	Cranial nerves (CNs) play essential roles in sensory, motor, and autonomic functions. Accurate CN parcellation from multimodal magnetic resonance imaging (MRI) is crucial for neuroanatomical analysis and neurosurgical planning. However, accurate CN parcellation remains extremely challenging because CNs are very small, exhibit low image contrast, and have slender tubular morphologies and complex anatomical trajectories. Moreover, the lack of publicly available, expert-annotated datasets has impeded the development and fair benchmarking of learning-based CN analysis methods. 
	In this work, we introduce CNsEMD, an expert-annotated multimodal dataset for CN parcellation. It comprises data from 202 subjects acquired on 3T, 5T, and 7T MRI scanners. We further propose the projective hyperspherical manifold network (PHM-Net), which learns cross-modal representations by capturing angular relationships in a shared hyperspherical embedding space. Rather than performing multimodal fusion in Euclidean space, the proposed Hyperspherical cross-modal interaction (HCI) module enables bidirectional feature exchange between T1-weighted (T1w) and direction-encoded color (DEC) representations on a unit hypersphere. The Magnitude-preserving projective hyperspherical orientation representation (PHOR) captures the axial nature of DEC orientations while preserving diffusion magnitude. The hyperspherical prototype segmentation head (HPSH) further extends angular similarity to voxel-wise classification using normalized voxel embeddings and learnable class prototypes. Extensive experimental results on the CNsEMD dataset demonstrate the effectiveness of our PHM-Net against state-of-the-art methods.
	CNsEMD establishes a reproducible benchmark for multimodal CN imaging, while PHM-Net provides a geometry-consistent solution for CN parcellation across diverse MRI acquisitions.
\end{abstract}


\section{Introduction}
\label{sec:introduction}
\IEEEPARstart{C}{ranial} nerves (CNs) are critical neural pathways that mediate sensory, motor, and autonomic functions \cite{sultana2017mri,zolal2016comparison,jacquesson2019full}. Accurate CN parcellation is essential for neuroanatomical analysis, neurosurgical planning, and the diagnosis and treatment of CN disorders \cite{yoshino2016visualization,li2024tractography}. In clinical routine, experienced specialists identify CNs by integrating complementary information from multimodal magnetic resonance imaging (MRI). However, developing an accurate and robust automatic CN parcellation framework remains challenging because CNs are extremely small, exhibit slender tubular morphology, have low image contrast, and follow complex anatomical trajectories.

CN pathways can be delineated by filtering streamlines generated using diffusion MRI tractography. In this framework, specific CN pathways are reconstructed by extracting the corresponding streamlines from whole-brain tractograms \cite{behan2017comparison,jacquesson2019overcoming}. However, tractography-based methods usually require manually defined regions-of-interest (ROIs) and extensive expert interaction \cite{he2021comparison,huang2025unified}. The resulting workflow is labor-intensive and difficult to reproduce across centers. Atlas-based tractography methods have therefore been developed to automate CN identification from diffusion MRI \cite{xie2025automated,zhang2020creation,zeng2023automated}. Nevertheless, these methods still involve several processing stages, including diffusion modeling, fiber orientation distribution (FOD) estimation, tractography, and bundle identification. Errors introduced at any stage can propagate through the pipeline and reduce the accuracy of the final CN delineation.
\begin{figure}[t]
	\centering
	\includegraphics[width=0.48\textwidth]{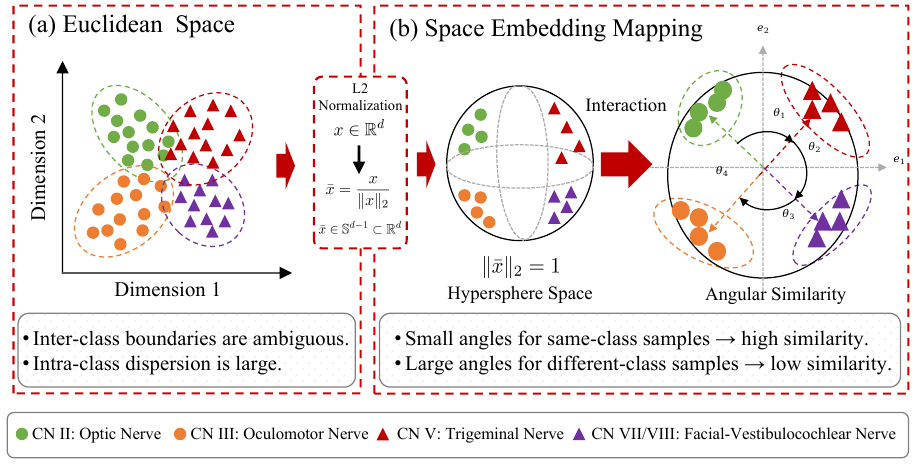}
	\caption{Conceptual motivation for hyperspherical cross-modal interaction. In Euclidean space, feature similarity depends on both direction and magnitude. HCI maps multimodal representations onto a unit hypersphere through $\ell_2$ normalization and models their cross-modal relationships using angular similarity.}
	\label{fig:1}
\end{figure}

Volumetric CN parcellation has emerged as an alternative approach by directly assigning each voxel to an anatomically defined nerve region based on structural MRI such as T1-weighted (T1w) and T2-weighted (T2w) images \cite{dolz2015fast,noble2011atlas}. This approach circumvents the complex streamline-generation pipeline entirely and enables true end-to-end voxel-wise CN delineation. Typical work is that Sultana et al. \cite{sultana2017mri} demonstrated direct CN parcellation using a deformable 3D contour model and surface representation derived from T2w images. However, accurate CN parcellation from a single structural modality remains inherently limited, as structural images alone provide insufficient contrast and lack directional information about fiber orientation. Multimodal learning addresses this limitation by jointly exploiting the complementary anatomical and directional cues available in structural and diffusion MRI.

Currently, multimodal CN parcellation networks integrating structural and diffusion MRI have achieved promising performance. The pioneering work, CNTSeg \cite{xie2023cntseg}, introduces a multimodal framework that directly segments major CNs from T1w, fractional anisotropy (FA), and FOD Peaks images. Building upon this, CNTSeg-v2 \cite{xie2025arbitrary} extends the paradigm to arbitrary-modal learning, improving adaptability to heterogeneous imaging protocols and accommodating a broader range of modalities, including T1w, T2w, FA, DEC, and Peaks images. Further advancing the CNTSeg series, DCLNet \cite{xie2025tractography} enforces anatomical consistency through tractography-guided dual-label supervision. Meanwhile, a diffusion-probabilistic model \cite{zhang2026dynamic} addresses the problem from a generative perspective, leveraging probabilistic diffusion representations to enhance CN localization. Despite these advances, two critical limitations persist.

\textit{Multimodal CN parcellation datasets.} Currently, there are no publicly available multimodal datasets with expert annotations for CN parcellation. Prior work has predominantly relied on private cohorts acquired with specific scanners and protocols, where variations in data sources, preprocessing pipelines, imaging modalities, and annotation criteria severely impede fair and reproducible comparison. The absence of a common benchmark further precludes systematic evaluation under heterogeneous imaging conditions. These factors underscore the pressing need for a public dataset that provides consistently annotated structural and diffusion MRI across diverse acquisition settings.

\textit{Fusion strategies for structural-diffusion images.} Existing methods fuse structural and diffusion features via concatenation, addition, or spatial attention, but largely treat diffusion representations as generic features and overlook their directional properties. Most fusion operates in Euclidean space, disregarding the distinct physical nature of each modality: T1w captures tissue anatomy, while DEC encodes principal diffusion orientation and magnitude. This geometric mismatch makes fusion sensitive to scale discrepancies and distribution shifts. Hyperspherical learning models directional relationships by $\ell_2$-normalizing features onto a unit hypersphere, where cross-modal similarity is measured by angle rather than magnitude \cite{sun2026multimodal,loshchilov2025ngpt,luo2026hyperspherical,xu2025deep} (Fig.~\ref{fig:1}). This provides a shared geometric space for T1w–DEC interaction, motivating our hyperspherical cross-modal interaction design.

In this paper, we present CNsEMD, an expert-annotated multi-field-strength dataset that combines structural and diffusion MRI from 202 subjects across 3T, 5T, and 7T. Furthermore, we propose PHM-Net, a projective hyperspherical manifold network for multimodal CN parcellation that operates in a shared hyperspherical embedding space, where cross-modal interaction is driven by angular similarity rather than magnitude. PHM-Net comprises three components: hyperspherical cross-modal interaction (HCI), which performs bidirectional angular feature exchange on the hypersphere; projective hyperspherical orientation representation (PHOR), which encodes DEC as a sign-invariant, magnitude-preserving orientation representation; and hyperspherical prototype segmentation head (HPSH), which classifies voxels by comparing normalized embeddings to learnable hyperspherical prototypes. Together, these modules form a geometrically consistent pipeline from encoding to parcellation. Extensive experiments on CNsEMD demonstrate that PHM-Net consistently outperforms state-of-the-art parcellation and multimodal learning methods, achieving superior accuracy and robustness across heterogeneous imaging conditions. 
Overall, the main contributions of this work are fourfold:

\begin{itemize}
	
	\item To our knowledge, CNsEMD is the first expert-annotated, multi-field-strength MRI dataset for multimodal CN parcellation, comprising 202 subjects scanned at 3T, 5T, and 7T and covering CN II, III, V, and VII/VIII. The dataset is available through the Science Data Bank\footnote{\url{https://doi.org/10.57760/sciencedb.44096}}, while the code and pretrained weights are available on GitHub\footnote{\url{https://github.com/IPIS-XieLei/CNsEMD}}.
	
	\item We propose PHM-Net, a projective hyperspherical representation learning for CN parcellation, which performs multimodal feature interaction and representation learning in a shared hyperspherical embedding space.
	
	\item We develop three components: HCI enables bidirectional feature exchange on a unit hypersphere through a shared angular affinity; PHOR learns a sign-invariant orientation representation while preserving diffusion magnitude; and HPSH classifies voxels by matching normalized embeddings to learnable hyperspherical prototypes.
	
	\item We evaluate the proposed PHM-Net against 12 state-of-the-art models on CNsEMD, encompassing general, multimodal, and CN-specific methods, and consistently achieve the best overall performance.
\end{itemize}

\section{CNsEMD Dataset}
\label{sec:dataset}
\subsection{Dataset Acquisition}
CNsEMD comprises 202 subjects acquired using 3T, 5T, and 7T MRI scanners, covering diverse imaging protocols, spatial resolutions, and image characteristics. For each subject, CNsEMD provides T1w images (obtained from the corresponding official websites of public datasets), DEC images reconstructed from diffusion MRI using MRtrix3~\cite{tournier2019mrtrix3}, and expert-curated voxel-level annotations. These labels cover five pairs of CNs: the optic (CN II), oculomotor (CN III), trigeminal (CN V), facial (CN VII), and vestibulocochlear (CN VIII) nerves. Because CN VII and CN VIII are closely apposed in the annotated region, they are treated as a single foreground category, yielding four foreground CN classes in total. All images were spatially aligned through a standardized preprocessing pipeline (diffusion processing, cross-modal registration, and spatial normalization). Table~\ref{tab:1} summarizes the characteristics of CNsEMD after preprocessing, while the original acquisition protocols of the three source datasets are described below.

HCP3T consists of 102 subjects randomly selected from the Human Connectome Project (HCP)~\cite{van2013wu}. The dataset provides T1w and multi-shell diffusion MRI acquired on a 3T Siemens Skyra scanner. Each diffusion acquisition contains 18 near-$b0$ images and 270 diffusion-weighted images distributed over three shells ($b=1000$, $2000$, and $3000~\mathrm{s/mm^2}$).

Diff5T consists of all 50 subjects from the publicly available Diff5T dataset~\cite{wang2025diff5t}. The dataset provides T1w and multi-shell diffusion MRI. Each diffusion acquisition contains 21 $b0$ images and 270 diffusion-weighted images distributed over three shells ($b=1000$, $2000$, and $3000~\mathrm{s/mm^2}$).

HCP7T consists of 50 subjects randomly selected from the HCP7T dataset~\cite{vu2015high}. The dataset provides high-resolution T1w and multi-shell diffusion MRI acquired on a 7T MRI scanner. Each diffusion acquisition contains 21 $b0$ images and 270 diffusion-weighted images distributed over three shells ($b=1000$, $2000$, and $3000~\mathrm{s/mm^2}$).

\begin{table}[h]
\centering
\caption{Summary of CNsEMD after preprocessing. Each subject contains one-channel T1w images and three-channel DEC images.}
\label{tab:1}
\setlength{\tabcolsep}{4pt}
\resizebox{0.48\textwidth}{!}{%
	\begin{tabular}{ccccc}
		\toprule[1pt]
		\textbf{Dataset} &
		\textbf{Field Strength} &
		\textbf{Matrix Size} &
		\textbf{Voxel Size} &
		\textbf{\#Cases} \\
		\midrule
		
		HCP3T
		& 3T
		& $145\times174\times145$
		& $1.25\times1.25\times1.25$
		& 102 \\
		
		Diff5T
		& 5T
		& $151\times181\times151$
		& $1.20\times1.20\times1.20$
		& 50 \\
		
		HCP7T
		& 7T
		& $173\times207\times173$
		& $1.05\times1.05\times1.05$
		& 50 \\
		
		\midrule
		
		\textbf{CNsEMD}
		& \textbf{3T/5T/7T}
		& --
		& \textbf{1.05--1.25}
		& \textbf{202} \\
		
		\bottomrule[1pt]
	\end{tabular}%
}
\end{table}
\begin{figure}[]
	\centering
	\includegraphics[width=0.48\textwidth]{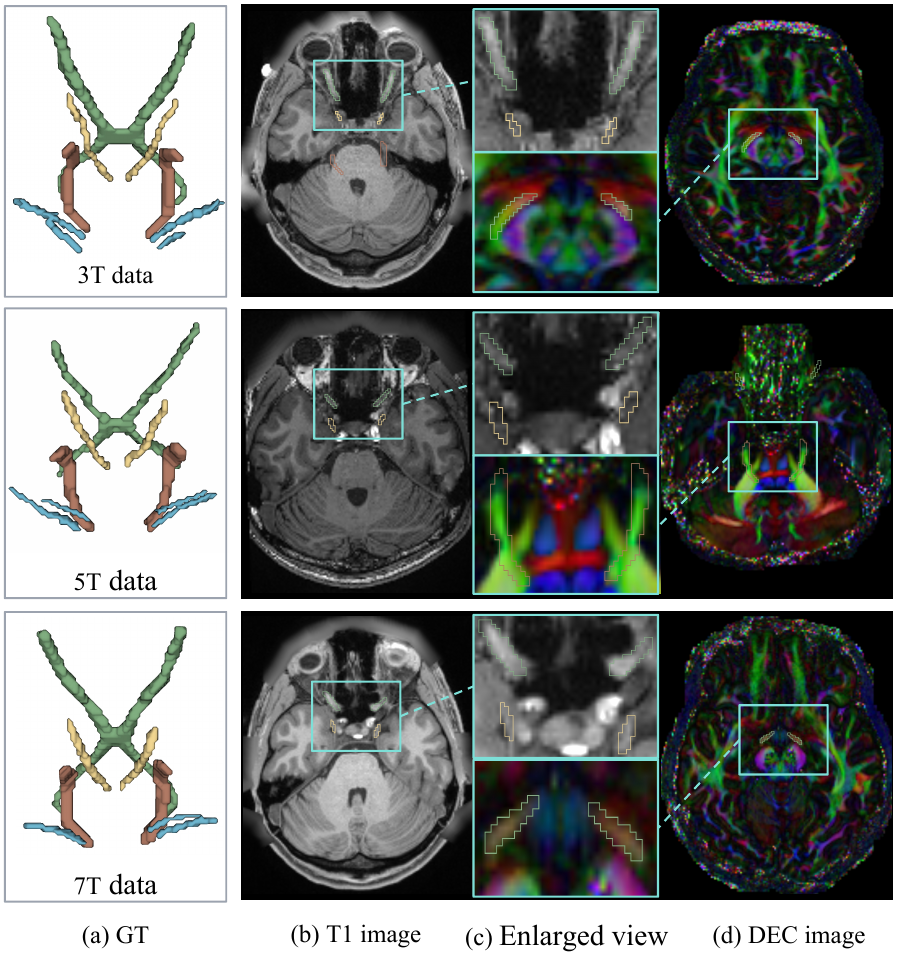}
	\caption{
		Representative CNsEMD cases grouped by the field strength of the
		diffusion acquisition (3T, 5T, or 7T). From
		left to right, each row shows the voxel-level annotations rendered
		as surface meshes, the T1w image, an enlarged view of the annotated
		CNs, and the corresponding DEC image.
	}
	\label{fig:dataset}
\end{figure}
\begin{figure*}[t]
	\centering
	\includegraphics[width=0.95\textwidth]{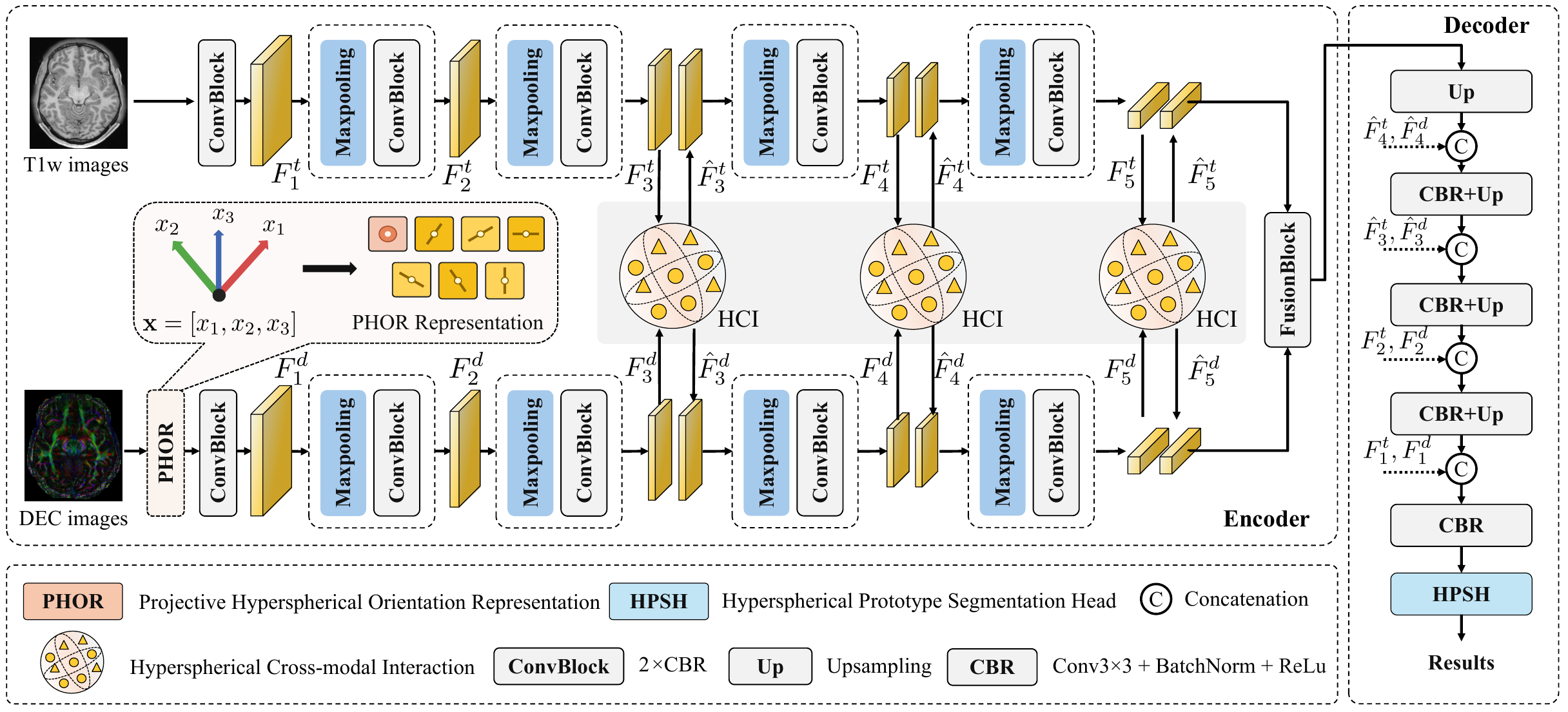}
	\caption{
		Overview of the proposed PHM-Net. PHOR first converts DEC vectors into sign-invariant, magnitude-preserving orientation representations. T1w and DEC features are then extracted by modality-specific encoders and progressively interacted through HCI in a shared hyperspherical space. The fused multiscale features are decoded, and HPSH produces the final voxel-wise parcellation based on angular similarities to learnable class prototypes.
	}
	\label{fig:2}
\end{figure*}
\subsection{Manual Annotation Protocol}
For the HCP3T subset, 102 cases were associated with reference annotations from our previous CNTSeg study \cite{xie2023cntseg}, where diffusion atlas guidance and multi-ROI selection were combined to delineate CN pathways. For the remaining Diff5T and HCP7T subsets, we employed a semi-automated strategy for voxel-level CN annotation. The annotation procedure was conducted as follows.
\textit{i)} The HCP3T reference annotations were used as the initial labeled set. To reduce the bias introduced by a single parcellation model, the HCP3T cases were divided into two subsets, denoted as $M_1$ and $M_2$, to train two independent nnUet models \cite{isensee2021nnu}, nnUNet$_1$ and nnUNet$_2$.
\textit{ii)} The trained nnUNet$_1$ and nnUNet$_2$ were applied to Diff5T and HCP7T to generate two preliminary CN masks, denoted as $M_{\mathrm{seg1}}$ and $M_{\mathrm{seg2}}$, respectively. The two predictions were merged to obtain the initial candidate masks $M_{\mathrm{init}}$ for subsequent manual refinement.
\textit{iii)} The candidate masks $M_{\mathrm{init}}$ from Diff5T and HCP7T, together with the HCP3T reference masks, were manually refined. Three medical postgraduate annotators corrected all masks slice by slice according to predefined anatomical criteria and multimodal MRI, producing refined masks $M_{\mathrm{refine}}$. During this step, intra-brainstem annotations were removed, and the masks were refined to improve anatomical plausibility, structural continuity, shape smoothness, and cross-dataset consistency.
\textit{iv)} Two experts with over five years of experience in CN anatomy further reviewed and corrected $M_{\mathrm{refine}}$, focusing on CN trajectories, anatomical continuity, and skull-base boundaries, to produce clinically refined masks $M_{\mathrm{double}}$.
\textit{v)} Finally, a senior neurosurgeon with more than ten years of clinical experience performed quality assessment of $M_{\mathrm{double}}$. Cases failing to satisfy the predefined annotation criteria were iteratively corrected until consensus was reached.
Through this multi-stage annotation procedure, voxel-level annotations were obtained for all 202 subjects. 
\section{Proposed Method}
\label{sec:METHODOLOGY}
\subsection{Overview of PHM-Net}

The overall architecture of the proposed PHM-Net is illustrated in Fig.~\ref{fig:2}. PHM-Net adopts a dual-branch encoder--decoder architecture that processes one-channel T1w images and three-channel DEC images. Before entering the diffusion branch, PHOR converts each DEC image into a seven-channel, magnitude-preserving representation invariant to the voxel-wise antipodal sign ambiguity of DEC orientations. The modality-specific encoders subsequently extract hierarchical anatomical and diffusion features. At the last three encoder stages ($l\in\{3,4,5\}$), HCI projects these features into a shared hyperspherical space and performs bidirectional cross-modal interaction based on angular similarity. A projection dimension of $D_h=512$ is used at all interaction stages. The updated bottleneck features are concatenated and processed by a convolutional fusion block. The decoder then progressively restores spatial resolution using multiscale skip features from both branches. Finally, HPSH classifies the normalized decoder embeddings according to their angular similarities to learnable class prototypes, producing the voxel-wise CN parcellation map. Together, PHOR, HCI, and HPSH form a coherent geometric pathway from DEC orientation encoding to cross-modal feature interaction and voxel-wise classification.

\subsection{Magnitude-Preserving Projective Hyperspherical Orientation Representation}

Diffusion MRI characterizes the direction-dependent diffusion of water in biological tissue. Under the diffusion tensor model, the voxel-wise diffusion tensor can be eigendecomposed as
\begin{equation}
	\mathbf{D}
	=
	\mathbf{V}
	\operatorname{diag}(\lambda_1,\lambda_2,\lambda_3)
	\mathbf{V}^{\mathrm T},
	\qquad
	\lambda_1 \geq \lambda_2 \geq \lambda_3,
\end{equation}
where $\mathbf{V}=[\mathbf{v}_1,\mathbf{v}_2,\mathbf{v}_3]$ contains the orthonormal eigenvectors, and $\operatorname{diag}(\lambda_1,\lambda_2,\lambda_3)$ is the diagonal matrix of the corresponding eigenvalues, with $\lambda_i$ associated with $\mathbf{v}_i$. The eigenvector $\mathbf{v}_1$ associated with the largest eigenvalue $\lambda_1$ represents the principal diffusion orientation, whereas FA quantifies the degree of diffusion anisotropy. A DEC map generated using MRtrix3 \cite{tournier2019mrtrix3} combines these quantities into the voxel-wise vector
\begin{equation}
	\mathbf{x}
	=
	\operatorname{FA}\,\mathbf{v}_1
	=
	[x_1,x_2,x_3]^{\mathrm T}.
\end{equation}
For visualization, the absolute values of the three Cartesian components are mapped to the red, green, and blue channels. However, treating the signed DEC components as ordinary image channels does not account for their sign ambiguity: $\mathbf{v}_1$ and $-\mathbf{v}_1$ represent the same physical diffusion orientation.
 Directly encoding the signed DEC components may therefore map these equivalent directions to disparate features. Conversely, normalizing the vector alone would discard the FA-related magnitude. PHOR addresses both limitations through a magnitude-preserving projective encoding applied before the DEC encoder. 
 
 For a DEC vector $\mathbf{x}=[x_1,x_2,x_3]^{\mathrm T}\in\mathbb{R}^{3}$, its magnitude $m$ and normalized orientation $\mathbf{d}$ are defined as
\begin{equation}
	m=\lVert\mathbf{x}\rVert_2,
	\qquad
	\mathbf{d}
	=
	\frac{\mathbf{x}}{\max(m,\epsilon)},
\end{equation}
where $\mathbf{d}=[d_1,d_2,d_3]^{\mathrm T}$ and $\epsilon$ prevents division by zero in voxels with negligible diffusion magnitude. We then embed the orientation using the six independent components of the symmetric outer product $\mathbf{d}\mathbf{d}^{\mathrm T}$:
\begin{equation}
	\boldsymbol{\phi}(\mathbf{d})
	=
	\left[
	d_1^2,\,
	d_2^2,\,
	d_3^2,\,
	\sqrt{2}d_1d_2,\,
	\sqrt{2}d_1d_3,\,
	\sqrt{2}d_2d_3
	\right]^{\mathrm T}.
\end{equation}
Because the outer product is invariant under sign reversal,
\begin{equation}
	\boldsymbol{\phi}(\mathbf{d})
	=
	\boldsymbol{\phi}(-\mathbf{d}),
\end{equation}
this mapping represents each orientation axis as an antipodal equivalence class in the quotient space
$\mathbb{S}^{2}/\{\mathbf{d}\sim-\mathbf{d}\}$, which is isomorphic to the real projective plane $\mathbb{R}\mathrm{P}^{2}$. The final PHOR representation is
\begin{equation}
	\boldsymbol{\Psi}(\mathbf{x})
	=
	\left[
	m,\,
	m\boldsymbol{\phi}(\mathbf{d})^{\mathrm T}
	\right]^{\mathrm T}
	\in\mathbb{R}^{7}.
\end{equation}
The first channel retains the magnitude $m$, while the six projective orientation channels are scaled by the same magnitude to preserve their association with diffusion anisotropy. PHOR is parameter-free and operates independently at each voxel, converting the original three-channel DEC image into a seven-channel representation.

\subsection{Hyperspherical Cross-modal Interaction (HCI)}

\begin{algorithm}[b]
	\caption{Hyperspherical cross-modal interaction at encoder stage $l$.}
	\label{alg:hci}
	
	\KwIn{T1w feature $F_l^t$ and DEC feature $F_l^d$}
	\KwOut{Interacted features $\hat F_l^t$ and $\hat F_l^d$}
	
	Project modality-specific features:
	$E_l^t\leftarrow\phi_l^t(F_l^t)$,
	$E_l^d\leftarrow\phi_l^d(F_l^d)$\;
	
	Flatten the spatial dimensions:
	$Z_l^t,Z_l^d\in\mathbb{R}^{B\times N_l\times D_h}$\;
	
	Map token embeddings onto the unit hypersphere:
	$S_l^t\leftarrow\operatorname{Norm}(Z_l^t)$,
	$S_l^d\leftarrow\operatorname{Norm}(Z_l^d)$\;
	
	Compute the angular affinity matrix:
	$A_l\leftarrow S_l^t(S_l^d)^{\mathrm T}$\;
	
	Compute bidirectional interaction weights:
	$P_l^{t\leftarrow d}\leftarrow
	\operatorname{Softmax}(A_l/\tau_h)$,
	$P_l^{d\leftarrow t}\leftarrow
	\operatorname{Softmax}(A_l^{\mathrm T}/\tau_h)$\;
	
	Aggregate features from the opposite modality:
	$\bar S_l^t\leftarrow P_l^{t\leftarrow d}S_l^d$,
	$\bar S_l^d\leftarrow P_l^{d\leftarrow t}S_l^t$\;
	
	Apply residual aggregation and hyperspherical normalization:
	$\hat S_l^t\leftarrow
	\operatorname{Norm}(S_l^t+\bar S_l^t)$,
	$\hat S_l^d\leftarrow
	\operatorname{Norm}(S_l^d+\bar S_l^d)$\;
	
	Reshape and reproject the interacted embeddings:
	$\hat F_l^t\leftarrow\rho_l^t(\hat S_l^t)$,
	$\hat F_l^d\leftarrow\rho_l^d(\hat S_l^d)$\;
	
	\Return{$\hat F_l^t,\hat F_l^d$}
	
\end{algorithm}
HCI establishes correspondence between heterogeneous T1w and DEC
representations according to their angular relationships.
T1w features primarily describe anatomical appearance, whereas DEC
features encode diffusion orientation and strength. Their channel
magnitudes are therefore not directly comparable in a Euclidean
feature space. HCI removes this scale dependence before measuring
cross-modal correspondence and preserves a symmetric information flow
between the two modality branches. The complete computational procedure is summarized in Algorithm~\ref{alg:hci}.

Let $F_l^t,F_l^d\in\mathbb{R}^{B\times C_l\times H_l\times W_l}$
denote the T1w and DEC feature maps at encoder stage
$l\in\{3,4,5\}$. Modality-specific projection functions first map
the features to a common dimension:
\begin{equation}
	E_l^t=\phi_l^t(F_l^t),
	\qquad
	E_l^d=\phi_l^d(F_l^d),
\end{equation}
where each projection comprises a $1\times1$ convolution, batch
normalization, and ReLU activation. After flattening the spatial
dimensions, the projected features are represented as
$Z_l^t,Z_l^d\in\mathbb{R}^{B\times N_l\times D_h}$, where
$N_l=H_lW_l$ and $D_h=512$. Each token is then $\ell_2$-normalized along the embedding dimension:
\begin{equation}
	S_l^t=\operatorname{Norm}(Z_l^t),
	\qquad
	S_l^d=\operatorname{Norm}(Z_l^d),
\end{equation}
where $\operatorname{Norm}(\mathbf{z})=
\mathbf{z}/\max(\|\mathbf{z}\|_2,\epsilon)$ is applied along the
embedding dimension.

The cross-modal angular affinity matrix is computed as
\begin{equation}
	A_l=S_l^t(S_l^d)^{\mathrm T}
	\in\mathbb{R}^{B\times N_l\times N_l}.
\end{equation}
For nonzero normalized embeddings, each element of $A_l$ corresponds
to the cosine similarity between a T1w token and a DEC token. Bidirectional
interaction weights are obtained using row-wise softmax normalization:
\begin{equation}
	P_l^{t\leftarrow d}
	=
	\operatorname{Softmax}\left(\frac{A_l}{\tau_h}\right),
	\qquad
	P_l^{d\leftarrow t}
	=
	\operatorname{Softmax}\left(\frac{A_l^{\mathrm T}}{\tau_h}\right),
\end{equation}
where $\tau_h$ controls the concentration of the interaction weights.
Smaller values of $\tau_h$ emphasize the most angularly consistent
cross-modal correspondences, whereas larger values yield more diffuse
feature aggregation. The two normalizations are performed separately,
allowing each modality to retrieve complementary information from the
other rather than imposing a single directional update.
The representations transferred from the opposite modality are
\begin{equation}
	\bar S_l^t=P_l^{t\leftarrow d}S_l^d,
	\qquad
	\bar S_l^d=P_l^{d\leftarrow t}S_l^t.
\end{equation}
They are combined with the original embeddings through residual
aggregation and renormalized onto the unit hypersphere:
\begin{equation}
	\hat S_l^t=
	\operatorname{Norm}\left(S_l^t+\bar S_l^t\right),
	\qquad
	\hat S_l^d=
	\operatorname{Norm}\left(S_l^d+\bar S_l^d\right).
\end{equation}
Finally, $\hat S_l^t$ and $\hat S_l^d$ are reshaped to their spatial
layouts and mapped back to $C_l$ channels using modality-specific
reprojection functions $\rho_l^t$ and $\rho_l^d$, each implemented by
a $1\times1$ convolution, batch normalization, and ReLU activation:
\begin{equation}
	\hat F_l^t=\rho_l^t(\hat S_l^t),
	\qquad
	\hat F_l^d=\rho_l^d(\hat S_l^d).
\end{equation}
The updated features are propagated to the next encoder stage and are
also used by the corresponding decoder skip connections. Applying HCI
at stages $3$--$5$ introduces interaction after progressively larger
receptive fields have been established. Thus, the decoder receives cross-modally refined features at multiple semantic scales, while the early high-resolution encoder stages remain unchanged.

\subsection{Hyperspherical Prototype Segmentation Head (HPSH)}

A conventional $1\times1$ convolutional classifier produces logits through unconstrained Euclidean inner products. It therefore does not explicitly preserve the angular geometry introduced by HCI during voxel-wise classification. In contrast, HPSH classifies each voxel according to the cosine similarities between its normalized decoder embedding and a set of learnable class prototypes. Each prototype represents a trainable class direction on the unit hypersphere. Predictions therefore depend on angular compatibility rather than the unconstrained magnitudes of the classifier weights and decoder features.

Let $F^{\mathrm{M}}\in\mathbb{R}^{B\times C\times H\times W}$ denote the fused multimodal feature map produced by the final decoder stage. An embedding function $g(\cdot)$ maps this feature map to a $D_p$-dimensional embedding space. It consists of a $3\times3$ convolution followed by batch normalization and ReLU activation, and a subsequent $1\times1$ convolution. The embedding at spatial location $i$ is normalized as
\begin{equation}
	\mathbf{z}_i
	=
	\operatorname{Norm}\left(g(F^{\mathrm{M}})_i\right),
	\qquad
	\mathbf{z}_i\in\mathbb{R}^{D_p},
\end{equation}
where $D_p=32$. For nonzero embeddings whose norms exceed $\epsilon$, this operation gives
$\mathbf{z}_i\in\mathbb{S}^{D_p-1}$.

For each parcellation class $c$, including the background class, HPSH maintains $K$ learnable prototypes
$\{\mathbf{p}_{c,k}\}_{k=1}^{K}$. These prototypes are normalized during each forward pass:
\begin{equation}
	\bar{\mathbf{p}}_{c,k}
	=
	\operatorname{Norm}\left(\mathbf{p}_{c,k}\right).
\end{equation}
The temperature-scaled cosine similarity between voxel $i$ and prototype $k$ of class $c$ is
\begin{equation}
	s_{i,c,k}
	=
	\frac{
		\mathbf{z}_i^{\mathrm T}\bar{\mathbf{p}}_{c,k}
	}{\tau_p},
\end{equation}
where $\tau_p$ denotes the prototype temperature. In the general multi-prototype setting, the class logits are obtained through log-mean-exp aggregation:
\begin{equation}
	o_{i,c}
	=
	\log\left(
	\frac{1}{K}
	\sum_{k=1}^{K}
	\exp\left(s_{i,c,k}\right)
	\right).
\end{equation}

The final model uses three prototypes per class ($K=3$). This multi-prototype formulation allows each class to capture heterogeneous voxel embeddings while retaining angular compatibility as the basis of the final voxel-wise decision.

\begin{table*}[htbp]
    \centering
    \caption{Comparative results on the mixed-field-strength CNsEMD subject-independent evaluation set and the 3T, 5T, and 7T subsets. Dice and Jac are reported in percentage (\%), while ASD and AHD are reported in millimeters (mm). Results are presented as mean$\pm$standard deviation. The best and second-best results are highlighted in red and blue, respectively.}
    \label{tab:methods_comparison}
    \resizebox{0.87\textwidth}{!}{%
    \begin{tabular}{c|c|cccc|cccc}
        \toprule[1pt]
        & \multirow{2}{*}{Methods} & \multicolumn{4}{c|}{Mixed ($n=40$)} & \multicolumn{4}{c}{3T subset ($n=20$)} \\
        & & Dice (\%) $\uparrow$ & Jac (\%) $\uparrow$ & ASD (mm) $\downarrow$ & AHD (mm) $\downarrow$ & Dice (\%) $\uparrow$ & Jac (\%) $\uparrow$ & ASD (mm) $\downarrow$ & AHD (mm) $\downarrow$ \\
        \hline
        \multirow{4}{*}{\begin{tabular}{@{}c@{}}Segmentation\\Methods\end{tabular}}
        & CSNet & 75.16\std{6.08} & 61.13\std{7.20} & 0.488\std{0.197} & 0.383\std{0.164} & 76.76\std{2.97} & 63.05\std{3.72} & 0.445\std{0.092} & 0.355\std{0.083} \\
        & MSUNet & 75.64\std{5.90} & 61.77\std{7.08} & 0.475\std{0.190} & 0.376\std{0.162} & \second{77.42}\std{3.06} & \second{63.95}\std{3.83} & 0.436\std{0.096} & 0.352\std{0.083} \\
        & SwinUNet & 67.82\std{6.90} & 52.44\std{7.26} & 0.922\std{0.587} & 0.742\std{0.432} & 70.73\std{4.15} & 55.74\std{4.68} & 0.693\std{0.201} & 0.572\std{0.173} \\
        & TransUNet & 74.64\std{5.65} & 60.44\std{6.80} & 0.486\std{0.150} & 0.386\std{0.131} & 75.94\std{3.14} & 62.05\std{3.85} & 0.461\std{0.093} & 0.373\std{0.080} \\
        \hline
        \multirow{4}{*}{\begin{tabular}{@{}c@{}}Multimodal\\Methods\end{tabular}}
        & FECCNet & 75.45\std{5.49} & 61.50\std{6.63} & 0.507\std{0.184} & 0.407\std{0.171} & 77.06\std{2.97} & 63.46\std{3.73} & 0.481\std{0.118} & 0.388\std{0.103} \\
        & NSYNet & 75.63\std{5.47} & 61.74\std{6.68} & 0.733\std{0.374} & 0.608\std{0.353} & 76.83\std{3.36} & 63.27\std{4.15} & 0.687\std{0.321} & 0.571\std{0.284} \\
        & F2Net & 70.71\std{5.62} & 55.62\std{6.40} & 0.567\std{0.176} & 0.454\std{0.152} & 71.98\std{3.18} & 57.12\std{3.58} & 0.534\std{0.125} & 0.438\std{0.105} \\
        & CMX & 65.07\std{5.84} & 49.23\std{6.11} & 0.746\std{0.357} & 0.604\std{0.258} & 65.83\std{3.98} & 50.18\std{4.05} & 0.701\std{0.141} & 0.588\std{0.124} \\
        \hline
        \multirow{5}{*}{\begin{tabular}{@{}c@{}}CN-specific\\Methods\end{tabular}}
        & MMFNet & 75.73\std{5.43} & 61.83\std{6.61} & 0.472\std{0.184} & 0.372\std{0.156} & 77.21\std{3.10} & 63.70\std{3.86} & \best{0.433}\std{0.090} & \best{0.345}\std{0.079} \\
        & DCLNet & 75.74\std{5.57} & 61.89\std{6.77} & 0.484\std{0.161} & 0.385\std{0.146} & 77.19\std{3.24} & 63.72\std{4.02} & 0.447\std{0.109} & 0.357\std{0.098} \\
        & CNTSeg & \second{75.90}\std{5.28} & \second{62.07}\std{6.39} & \second{0.455}\std{0.136} & \second{0.358}\std{0.122} & 77.26\std{3.16} & 63.74\std{3.96} & 0.435\std{0.081} & 0.349\std{0.071} \\
        & CNTSeg-v2 & 75.86\std{5.89} & \second{62.07}\std{7.15} & 0.536\std{0.318} & 0.432\std{0.284} & 77.29\std{3.41} & 63.86\std{4.23} & 0.474\std{0.133} & 0.385\std{0.114} \\
        & PHM-Net & \best{76.58}\std{5.33} & \best{62.92}\std{6.55} & \best{0.453}\std{0.134} & \best{0.358}\std{0.121} & \best{77.86}\std{3.31} & \best{64.55}\std{4.17} & \second{0.434}\std{0.094} & \second{0.346}\std{0.083} \\
        \midrule[1pt]
        & \multirow{2}{*}{Methods} & \multicolumn{4}{c|}{5T subset ($n=10$)} & \multicolumn{4}{c}{7T subset ($n=10$)} \\
        & & Dice (\%) $\uparrow$ & Jac (\%) $\uparrow$ & ASD (mm) $\downarrow$ & AHD (mm) $\downarrow$ & Dice (\%) $\uparrow$ & Jac (\%) $\uparrow$ & ASD (mm) $\downarrow$ & AHD (mm) $\downarrow$ \\
        \hline
        \multirow{4}{*}{\begin{tabular}{@{}c@{}}Segmentation\\Methods\end{tabular}}
        & CSNet & 66.96\std{5.40} & 51.24\std{5.63} & 0.723\std{0.236} & 0.584\std{0.173} & 80.16\std{1.85} & 67.16\std{2.55} & \second{0.337}\std{0.038} & 0.236\std{0.028} \\
        & MSUNet & 67.51\std{4.83} & 51.90\std{5.16} & 0.689\std{0.240} & 0.562\std{0.188} & 80.20\std{2.00} & 67.27\std{2.75} & 0.340\std{0.039} & 0.240\std{0.032} \\
        & SwinUNet & 59.28\std{6.90} & 43.21\std{6.20} & 1.583\std{0.835} & 1.259\std{0.558} & 70.55\std{3.29} & 55.05\std{3.87} & 0.718\std{0.139} & 0.563\std{0.134} \\
        & TransUNet & 67.03\std{4.24} & 51.18\std{4.56} & 0.664\std{0.149} & 0.542\std{0.115} & 79.63\std{1.95} & 66.48\std{2.71} & 0.359\std{0.040} & 0.256\std{0.034} \\
        \hline
        \multirow{4}{*}{\begin{tabular}{@{}c@{}}Multimodal\\Methods\end{tabular}}
        & FECCNet & 67.75\std{3.85} & 52.17\std{4.04} & 0.713\std{0.196} & 0.602\std{0.178} & 79.91\std{1.98} & 66.91\std{2.73} & 0.353\std{0.040} & 0.252\std{0.034} \\
        & NSYNet & 68.43\std{3.87} & 52.84\std{4.16} & 1.026\std{0.408} & 0.906\std{0.407} & 80.43\std{2.11} & 67.60\std{2.87} & 0.532\std{0.237} & 0.384\std{0.176} \\
        & F2Net & 63.27\std{4.26} & 47.01\std{4.43} & 0.767\std{0.173} & 0.627\std{0.131} & 75.63\std{2.24} & 61.24\std{2.87} & 0.431\std{0.049} & 0.311\std{0.045} \\
        & CMX & 58.43\std{5.19} & 41.95\std{4.93} & 1.049\std{0.570} & 0.846\std{0.359} & 70.20\std{2.53} & 54.62\std{2.91} & 0.532\std{0.067} & 0.395\std{0.064} \\
        \hline
        \multirow{5}{*}{\begin{tabular}{@{}c@{}}CN-specific\\Methods\end{tabular}}
        & MMFNet & 68.56\std{4.42} & 52.97\std{4.77} & 0.671\std{0.244} & 0.548\std{0.189} & 79.93\std{2.39} & 66.94\std{3.30} & 0.351\std{0.050} & 0.250\std{0.043} \\
        & DCLNet & 68.33\std{4.46} & 52.71\std{4.75} & 0.669\std{0.165} & 0.557\std{0.135} & 80.26\std{1.93} & 67.39\std{2.66} & 0.374\std{0.064} & 0.268\std{0.054} \\
        & CNTSeg & \second{68.84}\std{4.12} & \second{53.46}\std{4.45} & \second{0.612}\std{0.146} & \second{0.500}\std{0.117} & 80.24\std{1.54} & 67.35\std{2.14} & 0.339\std{0.024} & \second{0.236}\std{0.023} \\
        & CNTSeg-v2 & 68.08\std{4.74} & 52.44\std{5.05} & 0.859\std{0.463} & 0.725\std{0.407} & \second{80.79}\std{2.00} & \second{68.11}\std{2.80} & \best{0.336}\std{0.047} & \best{0.236}\std{0.038} \\
        & PHM-Net & \best{69.65}\std{4.21} & \best{54.28}\std{4.55} & \best{0.596}\std{0.134} & \best{0.489}\std{0.109} & \best{80.94}\std{1.88} & \best{68.31}\std{2.64} & 0.349\std{0.065} & 0.249\std{0.058} \\
        \bottomrule[1pt]
    \end{tabular}%
    }
\end{table*}
\begin{figure*}[h]
	\centering
	\includegraphics[width=0.86\textwidth]{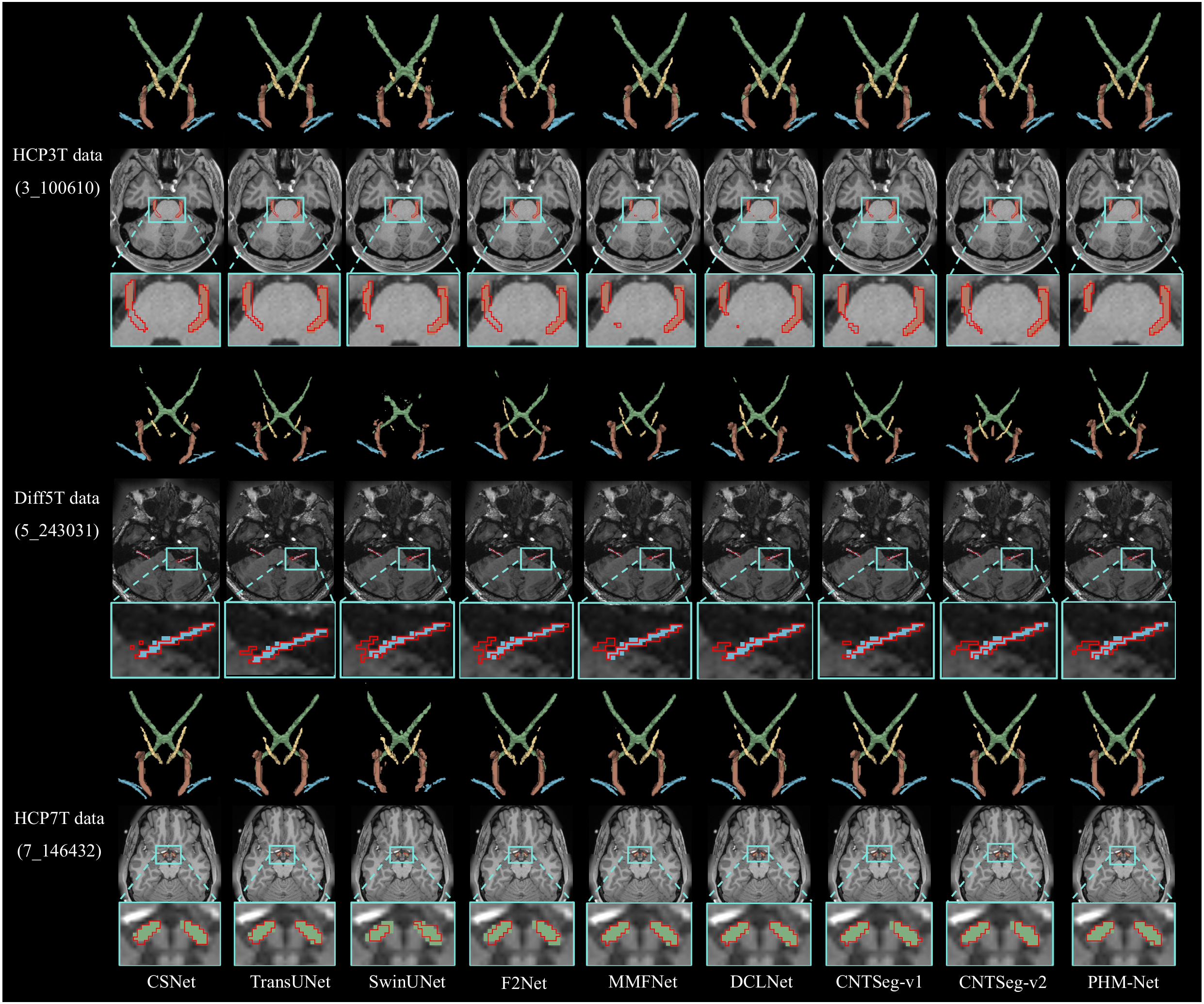}
\caption{Qualitative comparison on representative 3T, 5T, and 7T CNsEMD cases. For each case, the rows show three-dimensional reconstructions, slice-wise results, and enlarged views of challenging regions. In the slice-wise views, colored voxel overlays indicate the expert reference, while red contours delineate each method's predictions.}
	\label{fig:vision1}
\end{figure*}

\subsection{Training Objective}
\label{sec:training_objective}

PHM-Net is trained end to end using a combination of cross-entropy and soft Dice losses. Let $N$ denote the total number of pixels in a mini-batch, $C=5$ the number of classes including the background, $y_i\in\{0,\ldots,C-1\}$ the ground-truth label of pixel $i$, and $p_{i,c}$ the softmax probability that pixel $i$ belongs to class $c$. The cross-entropy loss is defined as
\begin{equation}
	\mathcal{L}_{\mathrm{CE}}
	=
	-\frac{1}{N}
	\sum_{i=1}^{N}
	\log p_{i,y_i}.
\end{equation}

To mitigate the severe imbalance between the background and the small CN structures, the soft Dice loss is computed over the four foreground classes:
\begin{equation}
	\mathcal{L}_{\mathrm{Dice}}
	=
	\frac{1}{C-1}
	\sum_{c=1}^{C-1}
	\left(
	1-
	\frac{
		2\sum_{i=1}^{N}p_{i,c}y_{i,c}+\epsilon
	}{
		\sum_{i=1}^{N}p_{i,c}
		+
		\sum_{i=1}^{N}y_{i,c}
		+
		\epsilon
	}
	\right),
\end{equation}
where $y_{i,c}\in\{0,1\}$ is the one-hot indicator of class $c$ and $\epsilon=10^{-5}$ ensures numerical stability. The overall training objective is defined as
\begin{equation}
	\mathcal{L}_{\mathrm{seg}}
	=
	\lambda_{\mathrm{CE}}\mathcal{L}_{\mathrm{CE}}
	+
	\lambda_{\mathrm{Dice}}\mathcal{L}_{\mathrm{Dice}},
\end{equation}
where $\lambda_{\mathrm{CE}}$ and $\lambda_{\mathrm{Dice}}$ control the relative contributions of the two terms. 

\section{Experiments}
\label{sec:EXPERIMENTS}
\subsection{Experimental Settings}

PHM-Net was implemented in PyTorch, and each fold was trained on a single NVIDIA RTX 5090 GPU. The loss weights in the training objective (Section~\ref{sec:training_objective}) were set to $\lambda_{\mathrm{CE}}=\lambda_{\mathrm{Dice}}=0.5$. We optimized the network using Adam with an initial learning rate of $2\times10^{-3}$, a batch size of 32, and a maximum of 200 epochs. The learning rate was halved whenever the mean validation Dice failed to improve for five consecutive epochs, and training was terminated after 20 consecutive epochs without improvement. A fixed random seed of 3407 was used in all experiments.

The network input consisted of paired one-channel T1w and three-channel DEC axial slices. All volumes were first center-cropped to $128\times160\times128$. Each T1w slice was independently clipped to its 1st--99th intensity percentiles and z-score normalized, while non-finite DEC values were set to zero and finite values were left unchanged. For training and validation, 162 of the 202 subjects were split into five folds, preserving the approximate $2{:}1{:}1$ ratio of 3T, 5T, and 7T subjects per fold. In each round, four folds trained and the remaining fold validated. Only slices with annotated CN voxels were used. No data augmentation was applied during training or validation. The checkpoint with the highest mean validation Dice was kept per fold. The remaining 40 subjects (20 3T, 10 5T, 10 7T) were reserved solely for testing. Test slices were processed without augmentation, and slice outputs were stacked into a 3D volume. For ensemble inference, logits from the five fold-specific models were averaged before voxel-wise label assignment, and the ensemble prediction was evaluated once per test subject.

For fair comparison, all methods used the same subject-level partitions, preprocessing pipeline, training schedule, and hardware environment. Single-stream methods received a four-channel input formed by concatenating the one-channel T1w and three-channel DEC images, whereas multimodal architectures received the two modalities through their respective branches. All output layers were configured to predict the background and four foreground CN classes. With the exception of CNTSeg-v2 and DCLNet, which were trained using their original training objectives, all compared methods were optimized with the parcellation objective defined in Section~\ref{sec:training_objective}. Performance was evaluated using the Dice, Jaccard index (Jac), average surface distance (ASD), and average Hausdorff distance (AHD)~\cite{wang2021annotation}. Metrics were averaged over four foreground classes covering five pairs of CNs and reported as mean~$\pm$~standard deviation across subjects.

\subsection{Comparison with State-of-the-Art Methods}
We compared PHM-Net against 12 representative methods spanning three categories. The general parcellation group consisted of CSNet \cite{mou2019cs}, MSUNet \cite{xie2022semi}, SwinUNet \cite{cao2022swin}, and TransUNet \cite{chen2021transunet}, which cover attention-based convolutional, multiscale, and Transformer architectures. The multimodal group comprised FECCNet, NSYNet \cite{avital2019neural}, F2Net \cite{yang2023flexible}, and CMX \cite{zhang2023cmx}, each employing distinct feature-fusion and cross-modal interaction mechanisms. The CN-specific group included MMFNet \cite{xie2023deep}, DCLNet \cite{xie2025tractography}, CNTSeg \cite{xie2023cntseg}, and CNTSeg-v2 \cite{xie2025arbitrary}. The above methods provide broad coverage of general-purpose segmentation, multimodal fusion, and task-specific CN parcellation architectures. Since CNTSeg-v2 supports arbitrary modality combinations through a non-dedicated training strategy, we evaluated it using the two modalities adopted in this study (T1w and DEC) as its full input set, without applying modality dropping. 

Table~\ref{tab:methods_comparison} reports the performance of PHM-Net and the compared methods on the mixed test cohort of 40 subjects. PHM-Net achieved the best mean performance across all four metrics, with a Dice score of 76.58\%, Jac of 62.92\%, ASD of 0.453~mm, and AHD of 0.358~mm. Compared with the recent CN-specific model CNTSeg-v2, PHM-Net improved Dice and Jac by 0.72 and 0.85 percentage points, respectively, while reducing ASD and AHD by 0.083 and 0.074~mm. It also outperformed the other CN-specific methods across all four metrics, supporting the effectiveness of the proposed hyperspherical representation framework. Notably, compared with the concatenation-based FECCNet using the same T1w and DEC inputs, PHM-Net improved Dice and Jac by 1.13 and 1.42 percentage points and reduced ASD and AHD by 0.054 and 0.050~mm, respectively. This comparison indicates that the proposed hyperspherical interaction more effectively exploits complementary DEC information than direct Euclidean concatenation, leading to improved multimodal representation alignment and anatomical delineation.

To further assess performance across field strengths, we grouped CNsEMD into 3T, 5T, and 7T subsets and evaluated PHM-Net on each subset separately. As shown in Table~\ref{tab:methods_comparison}, PHM-Net achieves the best Dice and Jac in all three subsets and the best values across all four metrics at 5T. Its ASD/AHD values are 0.434/0.346 at 3T, 0.596/0.489 at 5T, and 0.349/0.249 at 7T. PHM-Net ranks second on both surface-distance metrics at 3T, while several competing methods obtain lower surface distances at 7T. Nevertheless, PHM-Net maintains the strongest overlap performance across field strengths and remains competitive in boundary localization. 

To qualitatively assess PHM-Net, we selected representative cases from the 3T, 5T, and 7T subsets of CNsEMD. Fig.~\ref{fig:vision1} shows Subjects 3\_100610, 5\_243031, and 7\_146432, respectively, including three-dimensional CN reconstructions, slice-wise results, and enlarged views of challenging regions. In the slice-wise views, the colored voxel overlays indicate the expert reference, while the red contours delineate the predictions. PHM-Net yielded more complete three-dimensional reconstructions and fewer false-positive regions than the competing methods. For Subject 3\_100610, its CN V prediction in the cisternal region more closely matched the expert reference, with fewer false positives. For Subject 7\_146432, PHM-Net more consistently delineated CN II along the optic tract from the optic chiasm toward the lateral geniculate nucleus. These observations complement the quantitative results and indicate improved delineation of fine CN structures.
\begin{table}[t]
	\centering
	\setlength{\tabcolsep}{2.5pt}
	\caption{Ablation study of the proposed components on CN parcellation performance.}
	\label{tab:component_ablation}
	\resizebox{0.43\textwidth}{!}{%
		\begin{tabular}{cccc|cccc}
			\toprule[1pt]
			Base
			& PHOR
			& HCI
			& HPSH
			& Dice (\%) $\uparrow$
			& Jac (\%) $\uparrow$
			& ASD (mm) $\downarrow$
			& AHD (mm) $\downarrow$ \\
			\midrule
			
			\checkmark & -- & -- & --
			& 75.63\std{5.47}
			& 61.74\std{6.68}
			& 0.733\std{0.374}
			& 0.608\std{0.353} \\
			
			\checkmark & -- & \checkmark & --
			& 75.87\std{5.50}
			& 62.05\std{6.67}
			& 0.531\std{0.226}
			& 0.430\std{0.207} \\
			
			\checkmark & \checkmark & -- & --
			& 76.15\std{5.42}
			& 62.36\std{6.63}
			& 0.488\std{0.161}
			& 0.391\std{0.149} \\
			
			\checkmark & \checkmark & \checkmark & --
			& 76.08\std{5.50}
			& 62.26\std{6.77}
			& 0.595\std{0.312}
			& 0.489\std{0.298} \\
			
			\checkmark & \checkmark & \checkmark & \checkmark
			& \textbf{76.58}\std{5.33}
			& \textbf{62.92}\std{6.55}
			& \textbf{0.453}\std{0.134}
			& \textbf{0.358}\std{0.121} \\
			
			\bottomrule[1pt]
		\end{tabular}%
	}
\end{table}
\begin{figure}[t]
	\centering
	\includegraphics[width=0.44\textwidth]{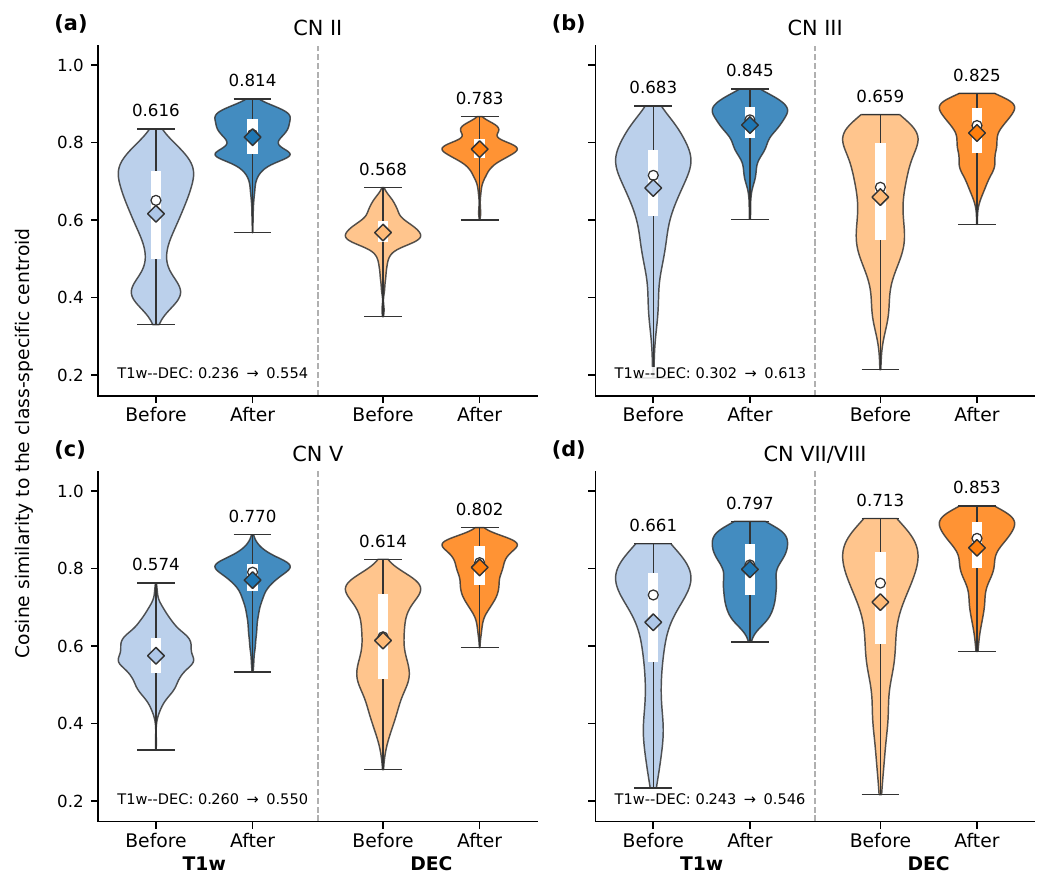}
	\caption{Modality-specific feature geometry before and after HCI at the third interaction stage. Each panel corresponds to one CN category, and its annotation reports the mean cosine similarity between spatially paired T1w and DEC embeddings.}
	\label{fig:hci_geometry}
\end{figure}
\subsection{Ablation Analysis}

PHOR, HCI, and HPSH were predefined before test-set evaluation, with all hyperparameters selected on the development cohort. The resulting ablation variants were then evaluated on the held-out test cohort without informing any architectural or hyperparameter decisions.

\subsubsection{Ablation of HCI}

As shown in Table~\ref{tab:component_ablation}, replacing Euclidean concatenation with HCI increased Dice from 75.63\% to 75.87\% and Jac from 61.74\% to 62.05\%, while decreasing ASD from 0.733 to 0.531~mm and AHD from 0.608 to 0.430~mm. The larger improvements in ASD and AHD indicate that HCI primarily improved boundary localization. Unlike direct concatenation, HCI measures angular similarity between normalized T1w and DEC features, allowing information to be exchanged with less influence from differences in feature magnitude. The feature analysis in Fig.~\ref{fig:hci_geometry} provides further evidence. Using the fold-0 PHM-Net checkpoint, we extracted embeddings from all 40 held-out test subjects and randomly sampled 1,500 foreground voxels per CN category with a fixed seed. Diamonds, circles, and white bars denote the mean, median, and interquartile range, respectively. HCI consistently increased the similarity of both T1w and DEC embeddings to their respective class-specific centroids and strengthened the correspondence between spatially paired cross-modal embeddings across all four CN categories. These results indicate improved within-modality compactness and cross-modal alignment.

\subsubsection{Ablation of PHOR and HPSH}

As shown in Table~\ref{tab:component_ablation}, adding PHOR to the Euclidean baseline increased Dice from 75.63\% to 76.15\% and Jac from 61.74\% to 62.36\%, while reducing ASD from 0.733 to 0.488~mm and AHD from 0.608 to 0.391~mm. This improvement supports the use of sign-invariant DEC orientation encoding. Combining PHOR with HCI achieved Dice and Jac scores of 76.08\% and 62.26\%, respectively, although ASD and AHD increased to 0.595 and 0.489~mm compared with the HCI-only model. After the conventional $1\times1$ classifier was replaced with HPSH, Dice and Jac increased to 76.58\% and 62.92\%, while ASD and AHD decreased to 0.453 and 0.358~mm. The complete model therefore achieved the best performance across all four metrics, supporting the role of prototype-based angular classification in translating the interacted hyperspherical features into anatomically accurate predictions.

\begin{figure}[t]
	\centering
	\includegraphics[width=0.46\textwidth]{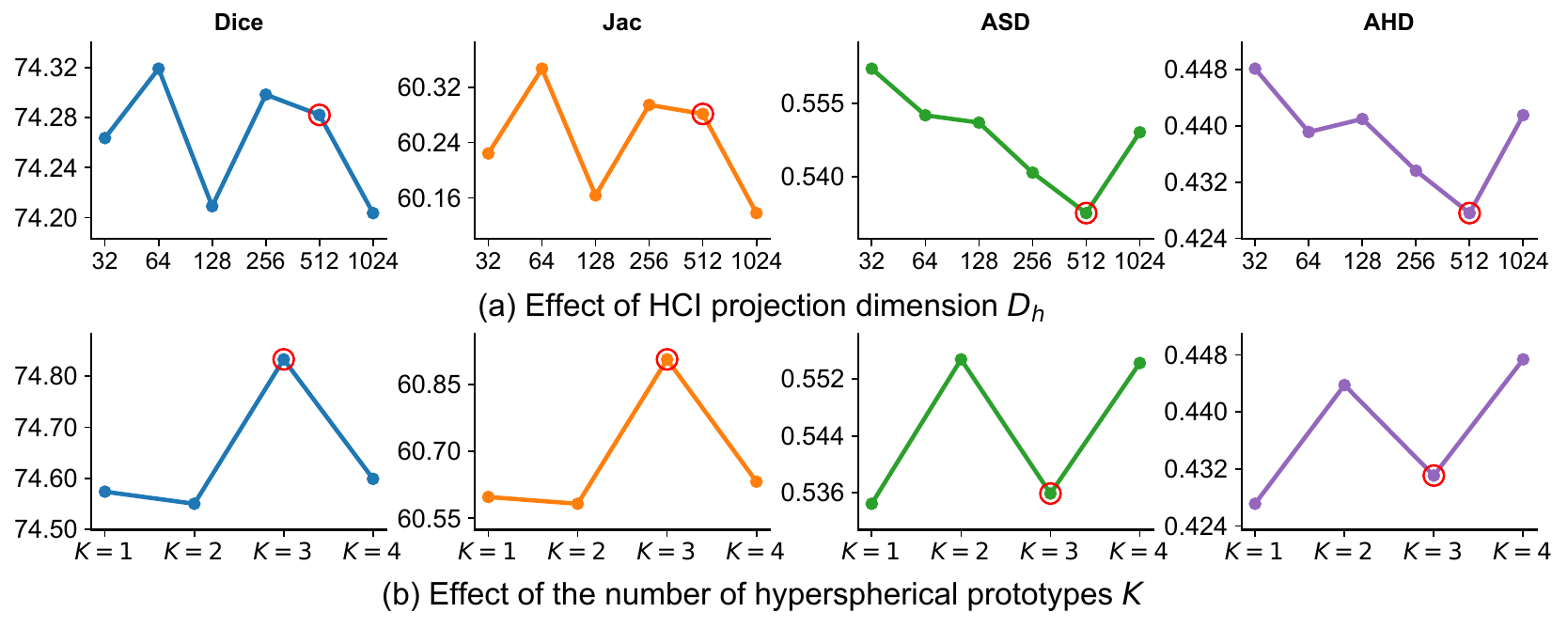}
	\caption{Parameter sensitivity analysis using five-fold out-of-fold validation results. (a) Effects of the HCI projection dimension $D_h\in\{32,64,128,256,512,1024\}$ on Dice, Jac, ASD, and AHD. (b) Effects of the number of hyperspherical prototypes $K\in\{1,2,3,4\}$ per class in HPSH. Red circles indicate the configurations adopted in the final model.}
	
	\label{fig:parameter_sensitivity_analysis}
\end{figure}
\begin{table}[t]
	\centering
	\setlength{\tabcolsep}{7pt}
	\caption{Computational complexity of the model variants. FLOPs are defined as twice the MACs reported by THOP for one $128\times160$ axial T1w--DEC input pair.}
	\label{tab:computational_complexity}
	\resizebox{0.4\textwidth}{!}{%
		\begin{tabular}{lcc}
			\toprule[1pt]
			Model variant & Params (M) & FLOPs/slice (G) \\
			\midrule
			Baseline                              & 81.34 & 55.43 \\
			Baseline + HCI                        & 85.03 & 57.80 \\
			Baseline + HCI + PHOR                 & 85.04 & 57.89 \\
			Baseline + HCI + PHOR + HPSH          & 85.06 & 58.68 \\
			\bottomrule[1pt]
		\end{tabular}%
	}
\end{table}
\begin{table}[t]
	\centering
	\setlength{\tabcolsep}{1.5pt}
	\caption{Performance comparison of different modality combinations for CN parcellation.}
	\label{tab:modality}
	\resizebox{0.39\textwidth}{!}{%
		\begin{tabular}{ccccc|cccc}
			\toprule[1pt]
			\textit{T1w} & \textit{T2w} & \textit{FA} & \textit{DEC} & \textit{Peaks}
			& Dice (\%) $\uparrow$
			& Jac (\%) $\uparrow$
			& ASD (mm) $\downarrow$
			& AHD (mm) $\downarrow$ \\
			\midrule
			
			\checkmark & & & & 
			& \textbf{71.83}\std{5.56}
			& \textbf{57.36}\std{6.22}
			& \textbf{0.587}\std{0.203}
			& \textbf{0.477}\std{0.183} \\
			
			& \checkmark & & &
			& 64.62\std{6.62}
			& 48.89\std{6.26}
			& 0.746\std{0.313}
			& 0.610\std{0.301} \\
			
			& & \checkmark & &
			& 58.82\std{7.70}
			& 42.78\std{7.25}
			& 1.030\std{0.731}
			& 0.836\std{0.490} \\
			
			& & & \checkmark &
			& 60.52\std{6.76}
			& 44.41\std{6.53}
			& 0.995\std{0.561}
			& 0.811\std{0.396} \\
			
			& & & & \checkmark
			& 55.95\std{11.58}
			& 40.38\std{10.31}
			& 1.064\std{0.492}
			& 0.890\std{0.417} \\
			
			\midrule
			
			\checkmark & \checkmark & & &
			& 72.75\std{6.05}
			& 58.40\std{6.57}
			& 0.602\std{0.493}
			& 0.485\std{0.402} \\
			
			\checkmark & & \checkmark & &
			& \textbf{73.91}\std{5.44}
			& \textbf{59.80}\std{6.22}
			& \textbf{0.558}\std{0.235}
			& \textbf{0.451}\std{0.204} \\
			
			\checkmark & & & \checkmark &
			& 73.67\std{5.00}
			& 59.53\std{5.73}
			& 0.567\std{0.228}
			& 0.454\std{0.190} \\
			
			\checkmark & & & & \checkmark
			& 73.08\std{5.70}
			& 58.93\std{6.48}
			& 0.592\std{0.234}
			& 0.472\std{0.197} \\
			
			\midrule
			
			\checkmark & \checkmark & \checkmark & &
			& 74.16\std{5.48}
			& 60.12\std{6.07}
			& 0.570\std{0.282}
			& 0.468\std{0.308} \\
			
			\checkmark & \checkmark & & \checkmark &
			& \textbf{74.33}\std{5.30}
			& \textbf{60.33}\std{5.94}
			& 0.571\std{0.249}
			& 0.459\std{0.211} \\
			
			\checkmark & \checkmark & & & \checkmark
			& 73.79\std{5.54}
			& 59.70\std{6.19}
			& 0.558\std{0.215}
			& 0.448\std{0.188} \\
			
			\checkmark & & \checkmark & \checkmark &
			& 73.78\std{4.95}
			& 59.57\std{5.80}
			& 0.568\std{0.184}
			& 0.458\std{0.166} \\
			
			\checkmark & & \checkmark & & \checkmark
			& 73.64\std{5.40}
			& 59.46\std{6.29}
			& \textbf{0.550}\std{0.194}
			& 0.446\std{0.175} \\
			
			\checkmark & & & \checkmark & \checkmark
			& 73.65\std{5.38}
			& 59.52\std{6.12}
			& 0.551\std{0.196}
			& \textbf{0.441}\std{0.173} \\
			
			\midrule
			
			\checkmark & \checkmark & \checkmark & \checkmark &
			& \textbf{74.59}\std{5.23}
			& 60.59\std{5.83}
			& \textbf{0.555}\std{0.226}
			& 0.447\std{0.204} \\
			
			\checkmark & \checkmark & \checkmark & & \checkmark
			& 74.57\std{5.43}
			& \textbf{60.63}\std{6.12}
			& \textbf{0.555}\std{0.250}
			& 0.448\std{0.229} \\
			
			\checkmark & \checkmark & & \checkmark & \checkmark
			& 74.18\std{5.31}
			& 60.09\std{5.87}
			& 0.561\std{0.224}
			& \textbf{0.446}\std{0.188} \\
			
			\checkmark & & \checkmark & \checkmark & \checkmark
			& 73.84\std{5.39}
			& 59.77\std{6.15}
			& 0.558\std{0.221}
			& 0.450\std{0.192} \\
			
			\midrule
			
			\checkmark & \checkmark & \checkmark & \checkmark & \checkmark
			& 74.43\std{5.32}
			& 60.43\std{5.83}
			& 0.569\std{0.515}
			& 0.444\std{0.308} \\
			
			\bottomrule[1pt]
		\end{tabular}%
	}
\end{table}
\subsection{Parameter Sensitivity Analysis}

Using five-fold out-of-fold validation on the 162-subject development cohort, we first evaluated the HCI projection dimension $D_h$ with the HCI-based model and subsequently evaluated the number of prototypes per class $K$ after incorporating PHOR and HPSH (Fig.~\ref{fig:parameter_sensitivity_analysis}). Dice and Jac remained relatively stable across the tested projection dimensions, showing that performance was not strongly affected by $D_h$ within this range. Although $D_h=64$ achieved the highest Dice and Jac, $D_h=512$ produced the lowest ASD and AHD while maintaining comparable overlap performance. We therefore selected $D_h=512$ based on its overall performance. For HPSH, $K=3$ achieved the highest Dice and Jac, while its ASD and AHD remained close to the best values obtained with $K=1$. By comparison, $K=2$ and $K=4$ showed weaker overall results. We therefore used three prototypes per class in the final model.

\subsection{Paired Statistical Comparison}

We compared PHM-Net with CNTSeg across 40 held-out test subjects using two-sided Wilcoxon signed-rank tests with Holm correction. PHM-Net significantly improved Dice and Jac, with median gains of 0.40 percentage points (95\% bootstrap CI: 0.07--1.14; adjusted $p$=0.0119) and 0.51 percentage points (95\% bootstrap CI: 0.09--1.43; adjusted $p$=0.0121), respectively. Differences in ASD and AHD were not significant (adjusted $p$=0.8555 for both).

\subsection{Computational Efficiency}

Although PHM-Net introduces hyperspherical cross-modal interaction and prototype-based segmentation, its computational overhead remains limited. The model contains 85.06M trainable parameters, only 4.57\% more than the baseline (81.34M), owing to the lightweight HCI and HPSH designs. HCI operates on compact $\ell_2$-normalized embeddings with a shared angular affinity, while the parameter-free PHOR adds only 0.09G FLOPs. PHM-Net requires approximately 10~s per subject for slice-wise inference and volume reconstruction. These results indicate a favorable trade-off between computational efficiency and parcellation accuracy, supporting practical multimodal CN parcellation.
\subsection{Input Modality Analysis}

Previous CN parcellation studies used different diffusion-derived inputs~\cite{diakite2025dual,avital2019neural,xie2023cntseg}, making it difficult to distinguish architectural improvements from differences in input information. To enable more consistent cross-method comparisons, we evaluated alternative modality combinations using the same concatenation-based FECCNet baseline through five-fold out-of-fold validation on the 162-subject development cohort. T1w provides anatomical contrast, whereas FA characterizes diffusion anisotropy without directional information. Peaks represent local fiber orientations, while DEC compactly encodes the principal diffusion orientation and FA-related magnitude. As shown in Table~\ref{tab:modality}, adding DEC to T1w increased Dice from 71.83\% to 73.67\% and reduced ASD from 0.587 to 0.567~mm. Although T1w+FA performed slightly better, the difference was small, and additional modalities produced only modest, non-monotonic gains. We therefore adopted T1w+DEC as a compact and competitive configuration that supports PHM-Net's orientation-aware design and propose it as a practical reference for future CN parcellation studies.

\section{Conclusion}

In this work, we introduced CNsEMD, an expert-annotated multimodal MRI dataset comprising 202 subjects acquired using 3T, 5T, and 7T scanners. We also proposed PHM-Net for multimodal CN parcellation in a shared hyperspherical embedding space. By incorporating sign-invariant DEC encoding, angular cross-modal interaction, and prototype-based classification, PHM-Net provides a unified geometry-consistent framework for multimodal representation learning and voxel-wise parcellation. Extensive experiments on CNsEMD show that PHM-Net achieves the best average performance among the compared parcellation and multimodal learning methods under heterogeneous imaging conditions.  Overall, this study provides a reproducible benchmark and an effective geometry-consistent solution for multimodal CN parcellation under heterogeneous MRI acquisition.

\begingroup
\renewcommand{\baselinestretch}{0.88}\selectfont
\bibliographystyle{IEEEtran}
\bibliography{myreference}
\endgroup

\end{document}